# Social Media for Mental Health: Data, Methods, and Findings

Nur Shazwani Kamarudin, Ghazaleh Beigi, Lydia Manikonda, and Huan Liu

School of Computing, Informatics and Decision Systems Engineering,
Arizona State University, Tempe AZ 85281, USA
{nurkamarudin,gbeigi,lmanikon,huanliu}@asu.edu

**Abstract.** There is an increasing number of virtual communities and forums available on the web. With social media, people can freely communicate and share their thoughts, ask personal questions, and seek peer-support, especially those with conditions that are highly stigmatized, without revealing personal identity. We study the state-of-the-art research methodologies and findings on mental health challenges like depression, anxiety, suicidal thoughts, from the pervasive use of social media data. We also discuss how these novel thinking and approaches can help to raise awareness of mental health issues in an unprecedented way. Specifically, this chapter describes linguistic, visual, and emotional indicators expressed in user disclosures. The main goal of this chapter is to show how this new source of data can be tapped to improve medical practice, provide timely support, and influence government or policymakers. In the context of social media for mental health issues, this chapter categorizes social media data used, introduces different deployed machine learning, feature engineering, natural language processing, and surveys methods and outlines directions for future research.

## 1 Introduction

Social media is a popular channel to spread information online. There are hundreds of millions of users that communicate with each other to share their thoughts, ideas, and personal experiences which overloads these channels with information. There are remarkable challenges when it comes to mental health issues [15]. A growing body of research have focused on understanding how social media activities can be use to analyze and improve the well-being of people, including mental health [41,35,21]. With the presence of social media data, it is now easier to study the trend of mental health problems and also to help researchers get information from social media to study mental health issues [15]. The easy access and use of social media allow users to update their social media profiles without time or space restriction [47]. This makes social media a preferable medium for researchers for their investigations. In addition, it is cost-effective for information seekers. Users can easily get health-related information [33].

## 1.1 Social Media and its Analysis

Social media has become a good source for data collection. There are different types of data that can be used from social media such as text, image, video, and audio. The amount of data on social media data increases rapidly. For example, on Twitter, 350,000 tweets are generated per minute and 500 million tweets are generated per day. A major factor that might affect social media users is the way they use social media because it can be very beneficial or toxic at the same time. For instance, active use of social media with two-way communication can be very beneficial to the user but it can also be destructive or toxic to the user [1].

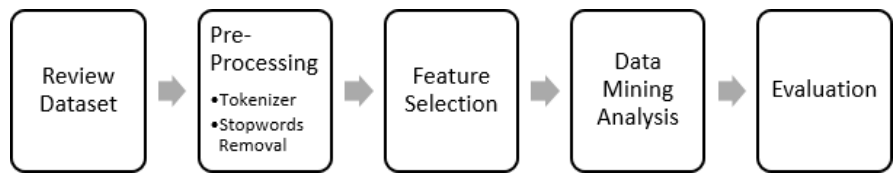

**Fig. 1.** Steps of social media analysis.

Figure 1 shows the common steps in social media analysis. It starts with dataset review in which the researchers need to choose the right dataset for their experiment. The second step is data pre-processing, which means preparing the data for the experiment such as removing stop words or word/sentence tokenizing. The next step is to select meaningful features from social media data such as an image or textual features. After selecting the right features, it is data mining analysis which includes deploying various techniques to develop the desired model. The final step is an evaluation, employing different metrics such as accuracy, recall, precision, F1 scores, for example.

## 1.2 Mental Health Problem

Mental health has become a public concern nowadays. People have started to think about the importance of mental health problems and their effects on our society. This is not a minor issue; on the contrary, it is a very serious issue that can contribute to mental well-being. For example, graduate students nowadays may face anxiety, depression, and stress because of the high competition in the academic world, long work hours, and lack support from their advisor [13,29]. These studies show that not just students [39,26,30,13,29], even employees [18,19,60] are all facing their own problems and authorities need to step up in order to help these groups of people. Researchers in the psychology field have studied this topic for decades. With the increasing use of social media data, this specific problem also attracts the attention of many computer scientists. Users actively share and communicate with online communities and researchers have found that it is a smart idea to leverage social media data to study this problem in order to help online communities and authorities at the same time. This can contribute to an immense change in overcoming this issue. Researchers have begun to investigate mental health problems such as post-traumatic stress disorder (PTSD) [34,16,48], depressive disorder [23,47,44,54], suicidal ideation [38,24,17], schizophrenia [42], and anxiety [58,22,28] through social media data.

This chapter is organized into five sections. Section 2 discusses the social media data on mental health by categorizing the data into three categories, which are linguistic-, visual- and combined- based. Section 3 presents a set of approaches used by the researcher on mental health research. In addition, evaluation metrics and findings are discussed in Section 4. Lastly, the study is concluded and a list of possible future works is presented in Section 5.

## 2 Social Media Data on Mental Health

Various types of social media data have been used by researchers to study mental health. Existing work could be categorized in to three groups based on the type of utilized social media data, 1) linguistic-based data, 2) visual-based data, and 3) combination of linguistic and visual data. Next, we will discuss each seperately.

### 2.1 Linguistic-based Data

Over the past few years, research in crisis informatics have utilized language as a medium to understand how major crisis events unfold in affected populations, and how they are covered on traditional media as well as online media such as blogs and social media sites [57]. Studies have shown that social media can provide a comforting environment for support seekers especially when it comes to stigmatized issues that make them reluctant to share with individuals around them.

Social media has been accordingly used to understand users' mental health issues. Interesting work by Coppersmith et al. [17] studies suicide attempts or even ideation among users using their posts in Twitter. The authors crawl tweets from 30 geolocations from all over the United States with at least 100 tweets per location. Then, they use natural language processing techniques to compare behavior of users who attempted suicide with those users who have previously stated that they were diagnosed with depression and neurotypical controls. In another work, Park et al. [47] focus on studying the impact of online social networks (OSN) on the depression issue. Accordingly, they collect two different kinds of data: 1) data from Internet-based screening test, which includes information of 69 participants who were asked to complete a questionnaire including depression related questions, and 2) collected tweets from Twitter from June 2009 to July 2009 that include the keyword 'depression'. After qualitative and quantitative analysis of collected data, authors show that social media data could be used to understand users' mental health issues.

Nadeem at al. [44] use social media data to study Major Depressive Disorder (MDD) issue in individuals. They use a publicly available dataset that was built from Shared Task organizers of Computational Linguistics and Clinical Psychology (CLPsych 2015) [16]. This dataset includes information of Twitter users who were diagnosed with depression. In another work, Amir et al. use Twitter to study depression and post-traumatic stress disorder (PTSD) [3]. They investigate the correlation between users' posts and their mental state. In particular, they investigate if tweets could be used to predict whether a user is affected by depression and PTSD or not. De Choudhury et al. [23] also show that social media can be utilized for predicting another mental health condition, i.e., major depressive disorder (MDD), using Twitter data. In this work, authors employ crowd-sourcing technique to provide the ground truth for their experiments. In another work [28], authors use Twitter data to study gender-based violence (GBV). The authors use Twitter's streaming API to sample a set of GBV related tweets based on a set of key phrases defined by the United Nations Population Fund (UNFPA) [52].

Another well-known social media platform is Reddit, which is a forum-based social media platform that captures communication between the original post and the user who left a comment on the thread. Each thread discusses a specific topic, which is known as a "subreddit". The work by De Choudhury et al. [22] uses Reddit to investigate how users seek mental health related information on online forums. They crawl mental health subreddits using Reddit's official API[1] and Python wrapper PRAW[2]. In another work, De Choudhury et al. [24] use Reddit to study the language style of comments left by users on the discussion forum in terms of influences towards suicidal ideation. This work fills the gap of how online social support can contribute to this specific problem. Authors use stratified propensity score analysis to determine if the user was affected by comments or not. They also estimate the likelihood that a user will receive a treatment based on the user's covariates. There was a study by [25] with work specifically on mental health subreddits such as r/depression, r/mentalhealth, r/bipolarreddit, r/ptsd, r/psychoticreddit. Based on the time stamp, data were divided into a treatment group and control group. These groups were further divided based on the causal analysis in order to analyze the effect of comments on the content of earlier posts with comment shared and received by users in their dataset.

Another work by Saha et al. [57] uses Reddit data to study the effect of gun violence on college students and the way they express their experience on social media. Authors collect related data from Reddit. Then they develop an inductive transfer learning approach [46] to see the pattern of stress expression by computing the mean accuracy value. In particular, they first build a classifier which labels the expressed stress in posts as *High Stress* and *Low Stress*. Then, they adopt the trained classifier to categorize collected posts from Reddit in order to identify posts that express higher stress level after the shooting incident.

---

[1] http://www.reddit.com/dev/api
[2] https://praw.readthedocs.org/en/ latest/index.html

Another work from Lin et al. [40] focuses on how online communities can affect the development of interaction within social media. This work investigates if new members will bring interruption in terms of perspectives towards social dynamic and lower content quality. The authors accordingly generate three questions related to user reception, discussion content and interaction patterns. They use Reddit data from Google BigQuery by choosing the top 10 subreddits between years 2013 and 2014. They study user reception, post content, and commenting patterns among Reddit users. This work studies the role of online social support based on historical data in conjunction with its effect on future health. It also investigates linguistic changes in online communities over time using data from two peer reviewing communities. Cohan et al. [14] study mental health by analyzing the content of forum posts based on the sign of self-harm thoughts. Their main goal was to study the impact of online forums on self-harm ideation. They consider 4 level of severity for the post content. Author build a model that includes lexical, psycholinguistic, contextual and topic modeling features. Their data were collected from a well-known mental health forum in Australia, namely, ReachOut.com[3].

## 2.2  Visual-based Data

The popularity of visual-based social media has increased rapidly. Users tend to communicate on social media by posting their photographs. Photo sharing provides a unique lens for understanding how people curate and express different dimensions of their personalities [4]. People use photos to define and record their identity, maintain relationships, curate and cultivate self-representation, and express themselves [62].

Posting pictures have become one way of communication among social media users. The definition of "selfie" is a picture that users take of themselves. Kim et al. [37] study the behavior of selfie-posting using Instagram data. This work uses selfie-posting to predict the intention of users who post selfies on social networking sites (SNSs). This work defines five hypotheses before they start designing the experiment. Those hypotheses were: attitude toward the behavior of selfie-posting, subjective norm, perceived behavioral control, and narcissism, which were possibly related to the intention to post selfies on social network sites (SNSs). The last hypothesis was the intention to post selfies on SNSs is positively related to the actual selfie-posting behavior on SNSs. They begin with 89 Instagram users. They recruit these users based on their agreement to be part of the study. Two coders analyze each user's account and the total sample size was ($n$ = 85). From the total number of participants, 9 were males and 76 were females. They also count the total number of the pictures posted on each user's account in a 6-week timestamp. Each user was required to answer a list of questions that were related to the standard Theory of Planned Behavior (TPB) variables such i.e., attitude, subjective norm, perceived behavioral control, and future intention based on [2].

---
[3] https://au.reachout.com/

Another work by Reece et al. [54] uses visual data for studying mental health in social media. The authors use Instagram data from 166 individuals with 43,950 photographs. In order to study the markers of depression, they use machine learning tools to categorize users to healthy and depressed groups. They began their experiment by crawling all posts from each user's account upon their agreement. Participant users were also required to answer a depression related questionnaire that contained specific questions based on inclusion criteria. In the last step of experiments crawled Instagram photos are rated using a crowd-sourced servic offered by Amazon's Mechanical Turk (AMT) workers.

### 2.3    Combined Data of Linguistic and Visual-based

Apart from using only visual data or only linguistic data, researchers also combine these two kinds of data to study social media influence on mental health. Sociologists also claim that it is not possible to communicate by using the only words; people also use pictures to communicate with each other [9]. A work by Burke et al. [11] used different features in Facebook data such as wall posts, comments, "likes", and consumption of friends' content, including status updates, photos, and friends' conversations with other friends in order to study the role of directed interaction between pairs. This work distinguish between two types of activity: directed communication and consumption. To do this, they recruited 1199 English-speaking adults from Facebook to be their research participants.

Andalibi et al. [4] study depression-related images from Instagram. They use image data and the matching captions to analyze if a user was having a depression problem or had faced this kind of problem in the past. They were keen on investigating if this group of users engage in a support network or not, and how social computing could be used to encourage this kind of support interaction among users. They gather 95,046 depression tagged photos posted by 24,920 unique users over one month (July 2014) using Instagram's API. All public details of each image were stored from these photos, such as user ID, number of likes and comments, date/time of creation, and tags. After conducting data collection, they begin the experiment by analyzing images and their textual captions. They also develop a codebook that includes 100 sample images and captions. Those coders then manually discuss the codebook in order to provide the best result for the experiment. Then, they add 100 more sample images and repeat the same steps.

Likewise, Peng et al. [50] investigate the effect of pets, relationship status, and having children, towards user happiness using Instagram pictures and captions. They use several hashtags such as #mydog, #mypuppy, #mydoggie, and #mycat, #mykitten, #mykitty to gather images of pet owner from Instagram. For non-pet owners, hashtags such as #selfies, #me, and #life were used to crawl the data. Before they started with the experimental steps, they began by classifying their data. The authors also provided the processed human face data called the face library for other researchers to use (see Table 1).

Manikonda et al. [41] use popular image-based media data in order to study mental health disclosure. This work extracts three main visual features from each image that they have in the corpus. Those features include visual features (e.g., color), themes, and emotions. Authors crawl the data from Instagram using Instagram's official API[4]. The main focus of this work was to study mental health disclosure based on visual features, emotional expression and how visual themes contrast with the language in a social media post. They specifically choose 10 mental health challenges from Instagram before they crawl two million public images and textual data from that particular medium. Those categories contain 10 types of disorders, which were: anxiety disorder, bipolar disorder, eating disorder, non-suicidal self-injury, depressive disorder, panic disorder, OCD, PTSD, suicide, and schizophrenia. Before they begin their experiment, they consult with the Diagnostic and Statistical Manual of Mental Health Disorders (DSM-V) in order to confirm that their final disorder categories could be reliable.

To sum-up, user-generated social media data is heterogeneous and consists of different aspects such as text, image, and link data. Table 1 summarizes different datasets used by researchers to study mental health using social media.

**Table 1.** List of mental health related social media data and available datasets.

| Type of Data | Paper | Name | Availability |
|---|---|---|---|
| Linguistic data | [44,17] | Computational Linguistic and Clinical Psychology (CLPsych) | Publicly available at https://bit.ly/2T0hMGO |
| | [57,40] | Google News dataset | Publicly available at https://bit.ly/1LHe5gU |
| | [47,44,28,22] | Reddit, Twitter, Facebook | Crawled using provided API by social media platform |
| Image data | [37] | Instagram | Publicly available at https://bit.ly/2SV5cbY |
| Combination of text and image data | [4,41] | Twitter, Facebook, Instagram | Crawled using provided API by social media platform |
| Link data | [23,36] | Twitter, Facebook | Crawled using provided API by social media platform |

## 3   Studying Mental Health in Social Media

As social media data has recently emerged as the main medium to spread information among online communities, there are also various approaches used by researchers to study related problems. This section discusses the approaches for studying mental health on social media. In this section, we elaborate on types of techniques or tools used in their reerarch. Figure 2 shows the categorization of social media analysis, namely machine learning methods, feature engineering and survey methods. Next, we introduce how social media analysis are used for mental health analysis in social media.

### 3.1   Machine Learning Methods

We discuss machine learning methods in terms of classification, clustering, and prediction with social media data for studying mental health problems.

---

[4] https://www.instagram.com/developer/

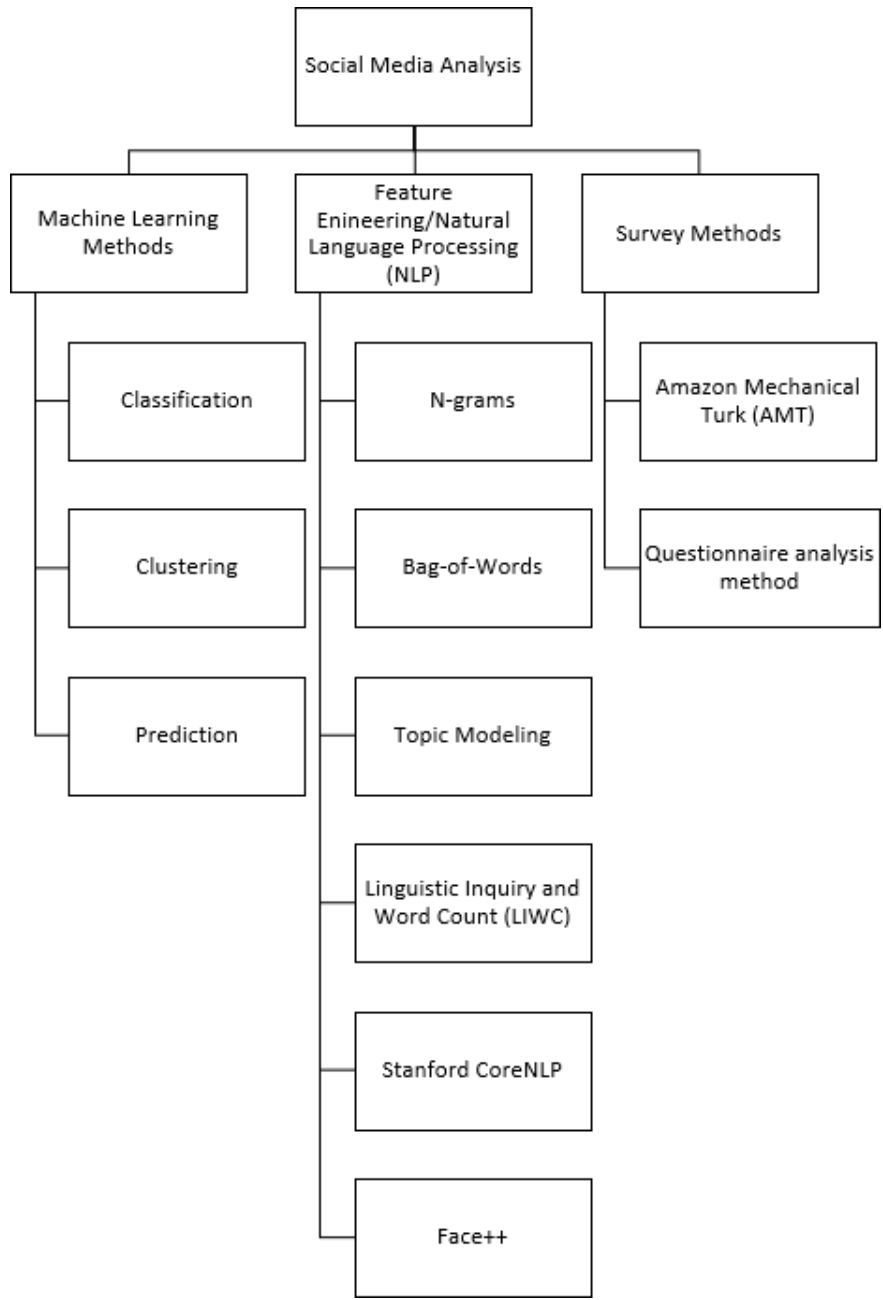

**Fig. 2.** Social media analysis approaches.

**Classification** In order to estimate the likelihood of having depression among users within a dataset, a work by Nadeem et al. [44] employ four types of classifiers (Decision Trees, Linear Support Vector Classifier, Logistic Regression, and Naive Bayes). They present a set of attributes to characterize the behavioral and linguistic differences of two classes. To do that, author utilize scikit-learn[5] which is a popular tool with many supervised and unsupervised machine learning algorithms [49].

**Clustering** De Choudhury et al. [23] cluster the ego-networks among users on social media. By clustering the ego-network, they study the characteristics of the graphs based on the egocentric measures, such as the number of followers, number of followees, reciprocity, prestige ratio, graph density, clustering coefficient, 2-hop neighborhood, embeddedness and number of ego components. Another work by [36] studies the correlation of social ties and mental health and finds that depressed individuals tend to cluster together.

**Prediction** The work by Reece et al. [54] predicts depression using photographic details, such as color analysis, metadata components and algorithmic face detection. In another approach, the authors in [23] divide users into two categories based on differences in behavior. For each user, they utilize a set of behavioral measures, such as mean frequency, variance, mean momentum, and entropy based on a user's one-year Twitter history. In order to avoid overfitting, the authors employ principal component analysis (PCA), then compared their method with several different parametric and non-parametric classifiers.

### 3.2 Feature Engineering/Natural Language Processing (NLP) Methods

Natural language processing plays a very important role in linguistic social media analysis. This subsection discusses feature representation techniques used in studying social media for mental health.

**N-grams** This text representational technique is widely adopted and is basically a set of co-occurring words within a given window. Features extracted using this technique are based on word frequency counts. In [22], authors calculate the most frequent unigrams from all Reddit posts and use negative binomial regression as their prediction model.

Saha et al. built a supervised machine learning model in order to classify stress expression in social media posts into binary labels of *High Stress* and *Low Stress* [57]. To develop the transfer learning framework for their experiment, they measure the linguistic equivalence by borrowing a technique from domain adaptation literature [20]. By using top $n$-grams ($n$ = 3) as an additional feature, Saha et al. developed a binary Support Vector Machine (SVM) classifier to detect *High Stress* and *Low Stress*. To build their training set, the authors extract 500 *n*-grams from the Reddit posts that they crawled. They compute the cosine similarity and compare their data with a Google News dataset in a 300-dimensional vector space. Authors find that it is possible to use social media content to detect psychological stress. On the other hand, as Twitter data has a limited number of characters per post, another work [10] designs a character *n*-gram language model (CLM) to get the score for each short text. This specific method examines sequences of characters including spaces, punctuation, and emoticons. For example, if we have a set of data from two classes, the model is trained by recognizing the sequence of characters. Similar character sequences will be classified into the same class. Given a novel text, the model can do estimations on which class can produce and generate all the texts.

---

[5] http://scikit-learn.org/stable/

Furthermore, the authors in [41] extract *n*-grams (*n*=3) to check the suitability and reliability of their corpus. Extracted *n*-grams have been further used to investigate if they are facing mental health disclosures. In order to extract visual features from the dataset, the authors pair OpenCV and Speeded Up Robust Features (SURF [7]). This approach is able to identify the meaningful themes from images. To study the linguistic emotions based on the visual themes, this work uses psycholinguistic lexicon LIWC and TwitterLDA. These two approaches help authors to measure the estimation of how themes and images were coherent to each other.

**Bag-of-Words (BOW)** Bag-of-Words (BOW) is a basic text representation for texts widely used by researchers in this area. When implementing this approach, a histogram is created to indicate how often a certain word is present in the text [63]. A previous work by [61] shows that the bag-of-words approach can be useful to identify depression. The author in [44] utilizes word occurrence frequencies to quantify the content from Twitter data by assembling all words and measuring the frequency of each word. Similarly, [14] uses BOW to extract features from their dataset. This work also uses content severity of the users in order to help forum moderators to identify the critical users that are keen on committing self-harm.

**Topic Modeling** One of the most mentioned topic modeling methods is Latent Dirichlet Allocation (LDA), which works by drawing distribution topics for each word in the document [8]. Then, words are grouped based on the distribution value. Similar words are in the same topic category. Cohan et al. [14] use the LDA model to find a set of topics from their data collection. By training the LDA topic model on the entire forum posts from their dataset, they are able to use the topic model as additional features for their experiment, which boosted the performance of their system and prove the effectiveness of topic modeling. Additionally, Manikonda et al. [41] use TwitterLDA to extract the linguistic themes from their dataset to see if visual and text are coherent to each other when it comes to mental health disclosure on Instagram.

Amir et al. [3] adopt a model known as Non-Linear Subspace Embedding (NLSE) approach [5] that can quantify user embedding based on Twitter post histories. The authors evaluate user embedding by using User2Vec (u2v), Paragraph2vec's PV-dm and PV-dbow models. They also leverage Skip-Gram in order to build vectors. Another design based on bag-of-embedding was bag-of-topics by using LDA to indicate topics presented in the user's posts. [22] They leverage LDA to identify types of social support on Reddit. They also consider information on practices that people share with the communities by characterizing self-disclosure in mental illness. The authors find that Reddit users discuss diverse topics. These discussions can be as simple as talking about daily routines but it can also turn into a serious discussion that involves queries on diagnosis and treatments.

Additionally, a work by [40] studies linguistic changes and for their data, the authors use several post-level measures including cross-entropy of posts and Jaccard self-similarity between adjacent posts. Then, the authors use the LDA model in order to compare the topic distribution among posts and general Reddit post samples. They also track the linguistic changes in sub-communities. In order to examine the interaction network's structural change, they calculate the exponent $a$ in the network's power-law degree distribution which gives the graph densification of the network [12]. Reddit allow users to vote on each post and comments, and the authors leverage this feature by computing the average score and complaint comment percentage in order to investigate community reaction to the content produced by newcomers.

**Linguistic Inquiry and Word Count (LIWC)**[6] It is a text analysis application that can be used to extract emotional attributes on mental health. This tool will be able to extract psycholinguistic features [14]. Manikonda et al. [41] use LIWC on texts associated with mental health images spanning different visual themes. LIWC can also characterize linguistic styles in posts from users [55]. Park et al. [47], use LIWC to quantify the level of depressive moods from their Twitter data. They compare a normal group vs. depressed group by measuring the average sentiment score from categories provided by the tools. LIWC contains a dictionary of several thousand words and each word fed to this tool will be scaled across six predefined categories: *social, affective, cognitive, perceptual, biological processes,* and *relativity*. Every criterion will have its own categories and sub-categories. For each sub-category, LIWC will assign specific scores for each word. The authors in [23] use LIWC to study Twitter users' emotional states. Then, they use point wise mutual information (PMI) and log-likelihood ratio (LLR) to extract more features from their corpus. Elsherief et al. [28] leverage LIWC in order to measure interpersonal awareness among users by differentiating perceived user and actual user characteristics.

A study by [22] captures the linguistic attributes of their data by measuring the unigram and then employ psycholinguistic lexicon LIWC. They choose

---

[6] http://www.liwc.net/

LIWC because it can categorize Redditors' emotions. They also examine the factors that drive social support on mental health Reddit communities, where the authors build a statistical model by measuring the top most frequent semantic categories from LIWC. The authors in [24] adopt the LIWC lexicon to study the various sociolinguistic features from their dataset and then measure the $t$-tests in order to analyze the differences between subpopulations. Coppersmith et al. [17] also use LIWC to study the pattern of language in conjunction with psychological categories generated from their dataset. This work uses LIWC to interpret how language from a given psychological category will be scored by the classifiers that they built. Likewise, Saha et al. [57] investigate on quantifying the psycholinguistic characterization. They employ LIWC measures to understand the psychological attributes in social media.

**Stanford CoreNLP** A study by [50] classifies the images and textual sentiments to see the significant role of those factors in reducing stress and loneliness among individuals. In order to interpret a user's happiness from a caption, they also utilize a sentiment analysis method, which is Valence Aware Dictionary and Sentiment Reasoner (VADER). Saha et al. [57] use Stanford CoreNLP's sentiment analysis model to retrieve the sentiment class of posts.

**Face++** A deep learning-based image analysis tool that is very useful for the facial recognition research field. It is an open-source face engine built with the convolutional neural network (CNN). In the work by [28], the authors use Twitter user's profile picture to predict the demographic information of the user by using Face++ API. Based on the GBV content, they investigate user involvement in GBV related post by leveraging the language nuances of those posts. Face++ is also used in [50] to do an experiment on face analysis by extracting user's information such as demographics inference, user relationship status, if a user has children, and then analyze user happiness.

### 3.3   Survey Methods

This subsection discusses works that use human intelligence tasks (HITs) for their analysis process, namely Amazon's Mechanical Turk and questionnaire based analysis. We also discuss the tools used.

**Amazon Mechanical Turk (AMT)** It is one of the widely used crowdsourcing platforms. On AMT, chunks of work are referred to as Human Intelligence Tasks (HIT) or micro-tasks [59]. This technique is leveraged in [22] to label words related to mental illness for their dataset. Moreover, Amazon's Mechanical Turk is used in [23] to take a standard clinical depression survey followed by several questions on depression history and demographics. The crowdworkers have the option to either include their *public* Twitter profiles or not in the analyzing process. Reece et al. use AMT service to rate the Instagram photographs collected for their experiment [54]. Raters are asked to judge how interesting, likable, happy, and sad each photo seemed, on a 0-5 scale.

**Questionnaire analysis method** The questionnaire can be very helpful for researchers to get more insight on the topic they are studying. For example, authors in [37] use this technique to measure attitude towards selfie-posting on semantically differential scales (e.g., bad/good, pleasant/unpleasant). By using the Narcissism Personality Inventory (NPI) [31], they measure Narcissism and participants' respond on a seven-point Likert scale (1 = "strongly disagree" to 7 = "strongly agree"). The authors uses AMOS 22 to test their hypotheses and see the relationships between attitude, subjective norm, perceived behavioral control, and narcissism toward their main question, which was the user's intention and behavior toward posting selfies on SNSs. In another work [54], authors uses the Center for Epidemiologic Studies Depression Scale (CES-D) [53] to screen participants' depression level for the depressed user group. Qualified participants are asked to share their Instagram usernames and history. An app embedded in the survey allow participants to securely log into their Instagram accounts and share their data.

Burke et al. [11] run a survey to analyze the relationship of social well-being and SNSs activity. Each user is required to answer survey questions using the format from the UCLA loneliness scale [56], Likert scales, and Facebook intensity scales [27]. They analyze their data without analyzing the private data of users, such as friend networks or identifiable information. They measure the number of friends and time spent on SNSs for each user so that they can get the answer to their research questions. In another work[47], authors use the CES-D to measure depressive symptoms and [23] also use CES-D to determine the depression levels of crowdworkers by distributing a depression survey to Amazon Mechanical Turk. As a summary, Table 2 shows the methods and tools that we discuss in this section.

**Table 2.** The summary of methods and tools in the application of mental health studies.

| Approach | Paper |
|---|---|
| Linguistic Inquiry and Word Count (LIWC) | [47,57,23,28,22,41] |
| Latent Dirichlet Allocation (LDA) | [3,22,41,8,14] |
| N-grams | [17,22,41,57,10] |
| Amazon Mechanical Turk (AMT) | [23,22,54] |
| Center for Epidemiologic Studies Depression Scale (CES-D) | [54,23,47] |

## 4  Evaluation Methods and Findings

In this section we discuss the utilized evaluation metrics and findings from the aforementioned papers. Fig. 3 represents the utilized evaluation metrics for a mental health studies in social media. We begin this section by overviewing evaluation metrics. Then, we discuss the findings from previous works.

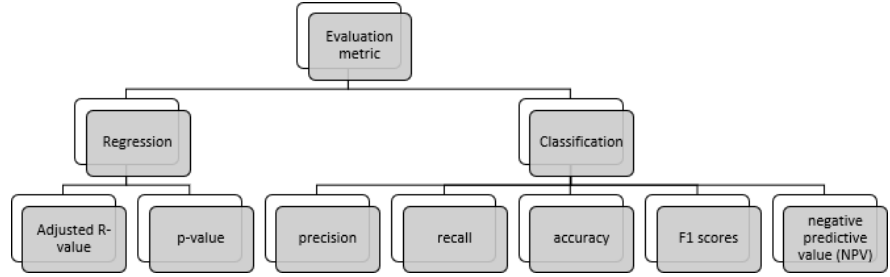

**Fig. 3.** Categories of evaluation metrics used for mental health analysis in social media.

### 4.1  Evaluation Metrics

There are various evaluation metrics in data mining analysis. We discuss the evaluation metrics used for studying mental health using social media data. The most common prediction metrics include precision, recall, F1 scores, and recipient operating classification (ROC) curves. Equation 1 shows the calculation of precision defined as the number of true positives (TP) divided by the sum of all positive predictions, TP and false positive (FP). Equation 2 is defined as the number of TP divided by the sum of all positives in the set, TP and false negative (FN). Equation 3 measures F1 scores by considering both precision and recall. The F1 score is the harmonic average of the precision and recall, and an F1 score reaches its best value at 1. Another metric known as adjusted R-squared is represented in equation 4. Adjusted R-squared is often used for explanatory purposes and explains how well the selected independent variable(s) explain the variability in the dependent variable(s). In adjusted R-squared, $n$ is the total number of observations and $k$ is the number of predictors. Adjusted R-squared is always less than or equal to R-squared.

$$precision = \frac{TP}{TP + FP} \quad (1)$$

$$recall = \frac{TP}{TP + FN} \quad (2)$$

$$F_1 = 2 \cdot \frac{precision \cdot recall}{precision + recall} \quad (3)$$

$$(R^2_{adj}) = 1 - [\frac{(1 - R^2)(n - 1)}{n - k - 1}] \quad (4)$$

Several works implement these metrics for their experiments [23,44,14,57]. De Choudhury et al. [23] evaluate their proposed classification approach by predicting individual depression level from their posts. They use precision, recall, accuracy, and receiver-operator characteristic (ROC) for evaluation. Their experimental results show that their classifier has a good performance in depression prediction. Another work from De Choudhury et al. [24] measures the most positive or negative $z$ scores in order to differentiate between mental health with influence risks to suicidal ideation (SW) users and mental health users. On the contrary, a study by [47] uses the coefficients from regression models to predict the Center for Epidemiologic Studies Depression Scale (CES-D) score. It then evaluates the proposed approach by measuring adjusted R-squared (equation 4) and p-value. Another work [3] measures *F1* and *binary F1* with respect to a mental condition in order to measure the performance of different models for its experiment.

Furthermore, authors in [28] measure the favorite rate and retweet rate for each tweet in order to count how many times a tweet was favorited and retweeted, respectively. These metrics are used to explore the engagement of users with gender-based violence (GBV) content on Twitter. Saha et al. [57] measure accuracy, precision, recall, F1-scores and ROC-AUC in order to see the performance level of their stress predictor classifier. Likewise, Coppersmith et al. [17] plot ROC curve of the performance for distinguishing people who attempted suicide from their age- and gender-matched controls. In order to compare the accuracy of all data and pre-diagnosis in their model prediction, [54] measure recall, specificity, precision, negative predictive value (NPV) and F1-scores.

Furthermore, a study by [41] calculates the Spearman rank correlation coefficients to compare the most frequent tags across all pairs of visual themes that belonged to six visual themes of mental health-related posts. Burke et al. present the ordinary least square (OLS) regressions for bridging and bonding social capital and loneliness based on the overall SNS activities [11]. Additionally, [40] study the effect of newcomers to existing online forums such as Reddit. They leverage the regression analysis by calculating the adjusted R-squared in order to measure the average score of post voting and complaint comment percentage on the content of the subreddit. In another work, Andalibi et al. calculate Cohen's Kappa coefficient to analyze depression related images along with the textual captions [4].

### 4.2 Output and Findings

In this section, we discuss the outputs and findings from previous works based on the type of data used. As we discussed in Section 2, we categorize works based on data that the authors used in their experiments. Here, we first discuss the findings of existing works that used linguistic data. Then, we review findings from visual data. Finally, we discuss findings from using combined data.

**Linguistic-based Findings** By focusing on linguistic-based experiments, [47] concludes that people disclose not only depressed feelings, but also very private and detailed information about themselves such as treatment history. For participants suffering from depression, their tweets were found to have high usage of words related to negative emotions and anger. In the end, users with depression tend to post more tweets about themselves than typical users. Likewise, the work by [23] demonstrates that Twitter can be used as a platform to measure major depression in individuals. To develop the prediction framework, the authors calculate four statistical values from the corpus including mean, variance, momentum, and entropy of selected features. Then, they compare these values between depression and non-depression classes. They find that individuals with low social activity tend to have a greater negative emotion, high self-attentional focus, increased relational and medicinal concerns, and heightened expression of religious thoughts. We can conclude that social activity does play an important role in individual mental well being. This group of people with low social activity has close-knit networks, which are normally highly embedded with their audience. The authors conclude that useful signals from social media can be used to characterize the onset of depression in persona by measuring their social activity and expression through their social networks. This kind of experiment shows how much social media can contribute to the body of knowledge in finding the solution and helping individuals in need.

Similarly, a work by [24] reports that comments play an important role in terms of giving support, especially among mental health communities. They observe that users who receive support from the online forums are more socially active and engage with the communities. Moreover, Nadeem et al. [44] show that Twitter can be used as a tool to predict MDD among users. The novelty of this research is the proposed text classification system which can classify if the tweets from users are depressive in nature or not. They conclude that social media can capture the individual's present state of mind. The text classification system is also effective because Twitter users are using this medium as a place to express their feelings. In addition, this study shows that it is reliable to use social media data for studying mental health related issues. Another study by Amir et al. [3] propose a novel model to extract users characteristics from their tweets known as user embeddings and further investigate their mental health status with respect to depression and PTSD. Their results show the correlation between captured embeddings and users' mental health condition.

In addition, Elsherief et al. [28] show that people discuss GBV-related issues on social medias. GBV-related hashtags help Twitter users to express their feelings, especially to share experiences and seek support. It has been shown that the most expressed emotion was anger. Another work demonstrates that communication between users plays an important role in mental well-being [14]. Their results show that when users are more active in communicating with other users, it helps to decrease the content severity. In another work, [40] compares the effects of growing online communities to the current network and showed that users perception remains positive and growth has an impact on users' attention. Authors also find that high levels of moderation helps to maintain the positive perception of community content after getting defaulted and that the communities' language do not become more generic or more similar to the rest of Reddit after the massive growth.

Saha et al. [57] show that posts published after gun-violence incident on college campuses include higher level of stress in comparison to those posted before incident. The authors also find that there is an increase in self-attention and social orientation when the campus population reduced. Also, more students were observed engaging in death-related conversations. In another work [17], researchers find that people who attempt suicide engage less in conversation, showing that this group of users had a smaller proportion of their tweets directed to other users. This work demonstrates how social media data can be beneficial for understanding mental health-related studies. De Choudhury et al. [22] also find that there is a variation in each type of social support. They also observe that posts related to self-attention, relationship and health issues received more attention from Reddit communities. Moreover, they find that negative posts get more attention compared to positive posts. By studying user feedback from all posts, they conclude that certain types of disclosures receives more social support from the online communities.

**Visual-based Findings**  Kim et al. [37] use the theory of planned behavior (TPB) [2] for studying behavioral intention using social media data. They find that all of their outlined hypotheses affect selfie-posting intention on social networks. Another work [54] extracts features from each images shared by users on social media platforms and shows that it is easy to distinguish between photos posted by healthy users and depressed ones. The results of this work demonstrate that healthy people will share photographs with higher hue value in comparison to depressed users who tend to share grayer images with lower brightness color. These results show that it is possible to detect depression through visual social media data. These findings confirm the fact that social media data could be used for mental health-related research.

**Combined-based Findings** A study by Andalibi et al. [4] find that Instagram users are aware of their audience. This is notable by observing how users address the audiences' concerns in captions of their posts. They find that images posted with a heavy amount of captions are related to support seeking and positive expression. Their results also show that specific hashtags on Instagram are used not only as semantic markers, but also as one way of categorizing content for the public. Similarly, Manikonda et al. [41] found that users used images to express their feelings such as emotional distress, and helplessness. Users' posts can be further used to understand how vulnerable and socially isolated they are. Results also show that images with various visual cues contribute to how users express themselves on Instagram. The authors finally conclude that Instagram is one of the mediums for users with mental health problems to seek help and also receive psychosocial support that they need. Another work by Burke et al. [11] demonstrates that direct communication affects users positively by making the user to bond with other users. This can further help users to feel less lonely. Their results also show that users with lower umber of interactions with others, tend to become more observer of other peoples' lives. In another work, Peng et al. [50] study the effect of owning a pet on personnel happiness by investigating posted images on social media. They compare users' happiness scores and find that pet owners were slightly happier than people who do not own pets. These results show the effectiveness of social media data in understanding users' behaviors and mental health related issues. In the next section, we summarize our findings and discuss potential future direction to expand research in this field.

## 5    Discussion and Future Directions

This chapter presents an overview of mental health related work that use social media data and machine learning in their studies. We discuss three key points focusing on mental health studies using social media data, namely, data, approaches, and findings. In the rest of this chapter, we provide a brief discussion followed by future work.

### 5.1    Discussion

Social media can affect users in many different ways. The main concern is if social media is beneficial to overcoming mental health problems among users. In [45], Facebook was rated as negative when it comes to cyber-bullying and bad sleeping patterns for users. But, when it comes to social support and building online communities, Facebook does help and was rated positively. Hence, it is important to make sure that social media be used in a good manner that can benefit users. Some significant effects of social media contributing to mental health well-being include: (1) social media can reduce stress in users through active communication with other users [14] and also provide information in capturing individual's present state of mind [44]; (2) social media is a popular channel for users to seek help and share information on the stigmatized issue. The anonymity of social media gives freedom to the user to express their feeling and might also improve in the stigmatized topic discussion [22]. (3) The use of social media with active communication may lead to improvement in the capability to share and understand others' feelings [11]. A study by Grieve et al. [32] indicates that Facebook connectedness may reduce depression and anxiety. Engaging with online communities can also give users the feeling of social appreciation through being understood [6].

On the other hand, some earlier studies point out the negative effects of social media to users. One finding is that users may have social isolation problems when they spend too much time on social media without having active communication with the online communities [23,17]. Increasing social media use without interaction with other users may cause depression, anxiety, sleep problems, eating disorder, and suicide risk. Primack et al. [51] report that in terms of subjective social isolation or perceived social isolation (PSI), increased time spent on social media can result in decreased traditional social experience, thus, increased social isolation, and exacerbate the feeling of exclusion.

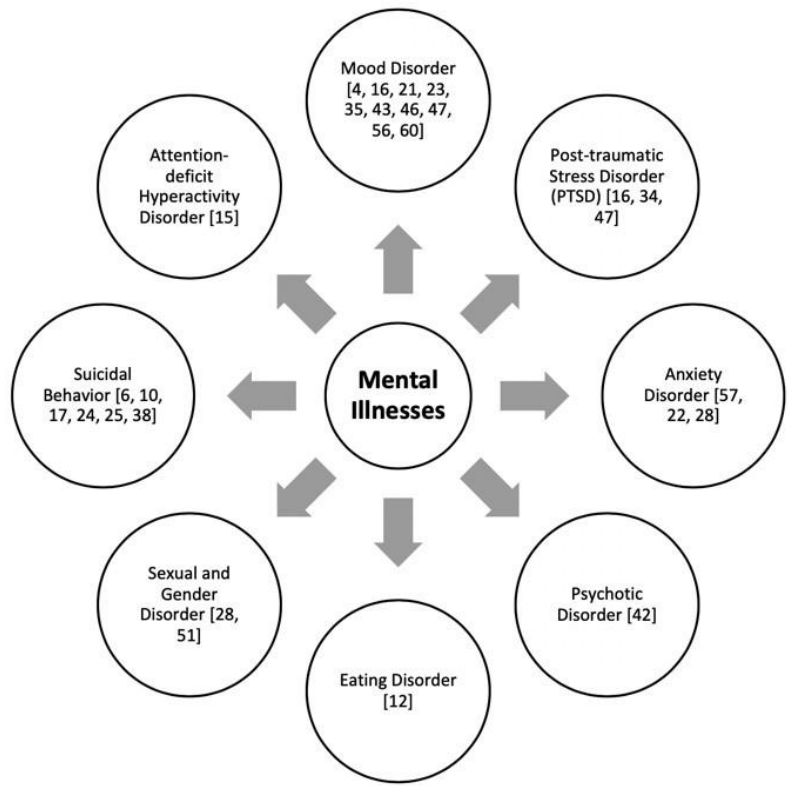

**Fig. 4.** An overview of different types of mental health issues and related work

Figure 4 shows different types of mental illnesses and related work using social media data and analysis. Existing studies are summarized in categories of mental health issues: (1) mood disorder [23,47,44,54], (2) post-traumatic stress disorder (PTSD) [34,16,48], (3) anxiety [58,22,28], (4) psychotic disorder [42], (5) eating disorder [12], (6) sexual and gender disorder [28], (7) suicidal behavior [38,24,17], and (8) attention deficit and hyperactivity disorder (ADHD) [15]. Mood disorder such as depression is studied in [16,21,61]. Depression is one of the mental health issues with high prevalence, receiving increasing attention lately. Limited work focuses on the psychotic disorder, eating disorder, and ADHD. These studies can also leverage social media data as it provides data about individuals' language and behavior [34]. Researchers use social media data to predict types of mental issues [37,23,21,12]. In [44] the severity of users' mental issues is estimated using social media data. A similar finding is reported in [42]. Network information available in social media data is leveraged for studying mental health issues [23,36].

### 5.2 Future Directions

Studying mental health issues using social media is challenging. Although a large body of work has emerged in recent years for investigating mental health issues using social media data, there are still open challenges for further investigation. Some potential research directions are suggested below:

- The increasing popularity of social media allows users to participate in online activities such as creating online profiles, interacting with other people, expressing opinions and emotions, sharing posts and various personal information. User-generated data on these platforms is rich in content and could reveal information regarding users' mental health situation. However, little attention has been paid on collecting the proper amount of user-information specifically on mental health [44,17]. One future direction is to collect a proper amount of labeled user-data as a benchmark which requires cooperation between psychologists and computer scientist [43]. This data can include users' behavioral information collected from social media platforms as well as their mental health condition information provided by experts. Preparing such data gives opportunities to both computer scientists and psychologists to benefit from a tremendous amount of data generated in social media platforms to better understand mental health issues and propose solutions to solve them.
- User-generated social media data is heterogeneous and consists of different aspects such as text, image, and link data. Most of the existing work investigates the mental health problem issues by just incorporating one aspect of social media data. For example, textual information is used in [57,17,47,44,16,3,23,28,22], image information is exploited [4,37,54], and link data in [23,36] to understand how user-generated information is correlated with people's mental health concerns. One potential research direction is to examine how different combinations of heterogeneous social media data (e.g., a combination of image and link data, combination of textual and link data, etc.) can be utilized to better understand people's behavior and mental health issues concern. Another future direction is to explore how findings from each aspect of social media data are different from each other, e.g., results w.r.t. textual data in comparison the findings w.r.t. link data.
- Most existing work utilizes either human-computer interaction techniques or data mining related techniques. For example, interview and surveys are used to help further study mental health related issues in social media [37,54,11,47,23,22]. Statistical and computational techniques are leveraged to understand users' behavior w.r.t. mental health issues [44,36,57,3]. However, research can be furthered to exploit both techniques to understand mental health issues in social media [23,54,47] and to develop both human-computer interaction and computational techniques specialized for understanding mental health issues for social media data.

- This chapter shows how different mental health issues have been studied using social media data. Fig. 4 represents different categories of mental health related issues using social media data. More mental health issues can be studied such as psychotic disorder, eating disorder, sexual and gender disorder.

Social media data can benefit mental health studies. The existing work shows that it is possible to study mental health by leveraging the large-scale social media data in understanding and analyzing mental health problems and employing machine learning algorithms to understand, measure, and predict mental health problems. More research on social media analysis using machine learning will help advance this important emerging field via multidisciplinary collaboration, research and development.

**Acknowledgements** We would like to thank all members of Data Mining Machine Learning Research Lab (DMML) at Arizona State University (ASU) for their constant support and feedback for this work. Special thanks to our lab members, Jundong Li, Matthew Davis, and Alex Nou for their detailed feedback on the earlier versions of this chapter. This work, in part, is supported by the Ministry of Higher Education Malaysia and University Malaysia Pahang (UMP).